\documentclass[runningheads]{llncs}
\usepackage{amsmath}
\usepackage{graphicx}
\usepackage{xcolor}
\usepackage{stmaryrd}
\usepackage{algorithm}
\usepackage[noend]{algpseudocode}
\usepackage{amssymb}
\usepackage{multirow}
\usepackage{adjustbox}
\usepackage{enumitem}
\usepackage{tikz}
\usetikzlibrary{arrows.meta, positioning, calc}

\title{Validating Generalist Robots with Situation Calculus and STL Falsification}
\author{Changwen Li\inst{1} \and Rongjie Yan\inst{1} \and Chih-Hong Cheng\inst{2} \and Jian Zhang\inst{1}}
\date{September 2025}
\institute{Key Laboratory of Software System, ISCAS, China
\and Carl von Ossietzky Universität Oldenburg, Germany}
\begin{document}

\maketitle

\begin{abstract}

Generalist robots are becoming a reality, capable of interpreting natural language instructions and executing diverse operations. However, their validation remains challenging because each task induces its own operational context and correctness specification, exceeding the assumptions of traditional validation methods. We propose a two-layer validation framework that combines abstract reasoning with concrete system falsification. At the abstract layer, situation calculus models the world and derives weakest preconditions, enabling constraint-aware combinatorial testing to systematically generate diverse, semantically valid world-task configurations with controllable coverage strength. At the concrete layer, these configurations are instantiated for simulation-based falsification with STL monitoring. Experiments on tabletop manipulation tasks show that our framework effectively uncovers failure cases in the NVIDIA GR00T controller, demonstrating its promise for validating general-purpose robot autonomy.

\end{abstract}

\section{Introduction}\label{sec:introduction}

Recent advances in robotic hardware, vision–language models, and accelerated computing are enabling general-purpose robot autonomy, with humanoid systems now capable of executing diverse natural-language tasks across domains from household assistance to disaster response~\cite{tong2024advancements}. Validating such robots is critical for trustworthiness and iterative improvement, yet substantially more challenging than in domains like autonomous driving. Whereas autonomous driving benefits from a well-defined operational design domain and fixed correctness criteria (e.g., reaching a destination while avoiding collisions), generalist robots must handle arbitrary tasks, each introducing its own applicable environment and correctness specification. Fully modeling their learning-enabled controllers is infeasible, so validation must rely on test generation that achieves both breadth (diverse operating conditions) and depth (bug hunting)~\cite{akazaki2018falsification,molin2023specification,zhang2024tolerance,li2023simulation}. This requires systematic and task-aware validation methods that reason over the operational domain and specification induced by each instruction, which is a capability that conventional falsification approaches lack.

To address the above challenge, we propose a two-layer validation framework. In particular, we introduce an abstract model of the robot system, whose purpose is to systematically generate configurations of initial world conditions and tasks to be performed. This abstract model focuses on (1) the structure of the world in which the robot operates, (2) the preconditions and effects of the abstract operations the robot may perform, and (3) the syntax and semantics of robot tasks formulated as programs over operations and the perceptual conditions available to the robot. We model the abstract robot system using Situation Calculus (SC)~\cite{reiter2001knowledge}, which captures evolving situations through first-order logical formulae, and we instantiate SC over finite domains to ensure decidability. Following our prior work on testing autonomous driving systems~\cite{li2022comopt}, the abstract model motivates us to perform combinatorial testing~\cite{nie2011survey} to create a relative completeness claim on the diversity of scenes and instantiated objects. As the SC formulation constrains the set of possible configurations (e.g., an apple can be in the cupboard but not vice versa), this leads to a revised constrained k-way coverage framework in which abstract scene instantiation should satisfy the constraints. In addition, we need to create diverse tasks for the robot to accomplish, with relative completeness claims established. For this, we consider bounded action steps in the task generation grammar and bounded cumulative depth of perceptual conditions, where k-way coverage is also used, and the feasibility of tasks is determined by computing weakest preconditions regressively.

Altogether, the first layer generates a set of world-task configurations that cover diverse situations and are logically guaranteed to be valid for completing the corresponding tasks. In the second layer, these configurations are executed on the real robot via mapping objects and properties in the abstract domain into the concrete domain (e.g., \textsf{closed-door} on cupboard object reflected as the angle between the door and the cupboard being less than $1^\circ$). We apply simulation-based falsification coupled with runtime verification. The specification of successfully completing the task is automatically synthesized from the task description into Signal Temporal Logic (STL)~\cite{maler2004monitoring,deshmukh2017robust} by applying the time-bounded eventual operator in a nested manner, characterizing the sequence of sub-task completion moments and enabling precise monitoring of the robot's behavior.

We have evaluated the proposed validation framework on the NVIDIA GR00T~\cite{bjorck2025gr00t} robot controller, controlling humanoid robots performing tabletop manipulation tasks, demonstrating that the abstract modeling effectively captures rich semantics and intertwined relations among objects in the environment, and that the validation approach is capable of uncovering a diverse range of defects. Altogether, this work enables a promising step toward performing systematic validation of generalist robots empowered by generative AI.

\section{Motivating Example}
We illustrate the challenge of validating generalist robots using a simple scenario with four objects (bread, plate, microwave, and table) and four primitive operations: putting one object into/on another, opening or closing a door, and turning on an appliance. For single-step tasks, executability and correctness are straightforward to determine. Admissible object pairs indicate when an action makes sense (for example, bread can be placed on a plate but not vice versa), and the expected effect offers a clear success criterion. This aligns well with existing simulation-based falsification approaches~\cite{nasiriany2024robocasa}.
Realistic tasks, however, involve sequencing, branching, and conditions. Correctness depends on the logical structure of the entire task and on inter-object dependencies. A sequence that begins with “close the microwave door” makes the subsequent “put the plate into the microwave” impossible. Conditional commands can be logically inconsistent (for example, “if no bread is detected, put the bread on the plate”). Operations also create implicit relational effects: placing a plate containing bread into the microwave implies the bread is inside too. Such cascading effects and global dependencies cannot be handled by per-operation admissibility rules alone, highlighting the need for a more expressive and situation-aware validation approach.

\section{Preliminaries}

In our robot validation framework, we alternate between two models that represent the same world at different granularities. Describing the abstract world model, tasks, and robot capabilities utilizes situation calculus (SC). The monitoring and identification of violating test cases are done by mapping a task in SC to the corresponding STL formula. 

\subsubsection*{Situation Calculus}  
It is a logical framework for modeling dynamic worlds~\cite{lin2008situation}.
SC is formulated over a sorted domain with three primary categories:  \textit{operations} that may be performed in the world,  \textit{situation} represents the history of operations applied from the initial situation, which is the state before any operation is executed, and \textit{objects} that denote all entities other than operations or situations. The evolution of \textit{world states} is described using \textit{fluents}, which capture properties of objects whose truth values may vary across situations. A world is specified in second-order logic through three classes of formulae: foundational axioms, formulae describing operations, including their preconditions and effects, and formulae describing properties of world states. A comprehensive introduction to SC can be found in~\cite{lin2008situation}. Second-order logic is in general undecidable. In Section~\ref{se:method}, we present a decidable variant of SC, obtained by restricting the domain of operations, situations, and objects to be finite.

\subsubsection*{Signal Temporal Logic Specification}

 The syntax of STL is defined as follows:
$\varphi ::= 
      \mu 
      \mid \lnot \varphi 
      \mid \varphi \land \psi 
      \mid \varphi \lor \psi 
      \mid \varphi \,\mathsf{U}_{[a,b]} \psi,
$
where $\mu$ is an atomic predicate of the form $f_\mu(\sigma(t)) > 0$, and $f_\mu$ is a function associated; and $\sigma(t)$ shows  the state of a finite execution trace of a system at time $t$; 
$\mathsf{U}_{[a,b]}$ is a \emph{bounded until} operator evaluated at time interval $[a,b]$. Intuitively, $\varphi \,\mathsf{U}_{[a,b]}\, \psi$ requires that $\varphi$ holds continuously until $\psi$ becomes true at some time in the interval $[a,b]$. 
The standard temporal operators \emph{eventually} and \emph{always} are also used, 
defined from $\mathsf{U}$: $
    \Diamond_{[a,b]} \varphi := \top \,\mathsf{U}_{[a,b]}\, \varphi,\;
    \Box_{[a,b]} \varphi := \lnot \Diamond_{[a,b]} \lnot \varphi.
$ 
The satisfaction of an STL formula by signals can be evaluated in a quantitative way using the \emph{robustness function} $\rho(\varphi, \sigma, t) \in \mathbb{R}$,
which measures the signed distance between the signal and the boundary of violation such that $\varphi$ is satisfied by $\sigma$ and $t$ only if $\rho(\varphi, \sigma, t) \ge 0$.  
The quantitative semantics of STL follows the standard recursive definition from the literature~\cite{deshmukh2017robust,maler2004monitoring}.

\section{Methodology}~\label{se:method}

\begin{figure}[t]
\center
\includegraphics[width=0.9\linewidth]{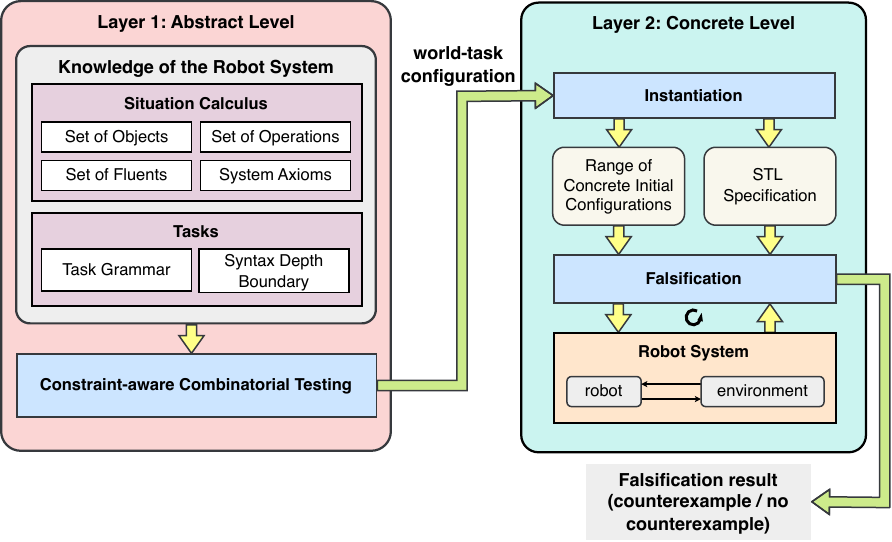}
\vspace{-3mm}
\caption{Validation Framework for Robots}
\label{fig:framework}
\vspace{-10pt}
\end{figure}

We present our two-layer validation framework (Fig.~\ref{fig:framework}). The first layer uses situation calculus to reason about the robot system and applies constraint-aware combinatorial testing to generate tasks and abstract world configurations. The second layer instantiates the configurations on the concrete system, formulates the corresponding falsification problems, and solves them to obtain counterexamples in which the robot fails to complete the intended task. 

\subsection{Characterizing Robot System with Situation Calculus}
\label{se:knowledge}

To enable tractable reasoning, we introduce an abstract model of the robot formulated in SC, which provides a structured representation for describing operations, their preconditions, and their effects. The abstraction captures objects and their qualitative relations, and specifies how these relations evolve when operations are executed according to system axioms. 
To ensure decidability, we consider a bounded SC with a finite domain of operations, objects, fluents, and situations.

\paragraph{Abstract Robot System} We present the formalization of SC in the context of a robot system. The objects represent physical entities relevant to the domain, such as the bread, plate, and microwave in the motivating example. Formally, these objects form a finite set $\mathcal O = \{o_1,...,o_n\}$ indicating  $n$ relevant objects in the world. Operations performed by the robot are parameterized by a tuple of these objects, $\alpha(\vec{o})$, such as $\textsf{put}(o_1, o_2)$ is informally interpreted as  ``put $\langle o_1\rangle$ into $\langle o_2\rangle$''. Situations, as the history of operations applied from the initial situation $s_0$, evolve inductively by $s_{i+1} = do(\alpha_i, s_i)$, where $do$ denotes the successor situation obtained by applying action $\alpha_i$ in situation $s_i$.

The state and the semantics of objects are represented using a finite set of predicates, which fall into two categories: \emph{rigid predicates}, denoted by $P(\vec{o})$, whose truth values remain invariant across all situations, and \emph{fluent predicates}, denoted by $F(\vec{o}, s)$, whose truth values may change from one situation to another. For example, a rigid predicate such as $\textsf{HasDoor}(o)$ indicates that $o$ has a door. A fluent predicate such as $\textsf{Loc}(o, o', s)$ states that $o$ is located in $o'$ in situation $s$, and its truth value may differ from that of $\textsf{Loc}(o, o', s')$ with another situation $s' \neq s$. In particular, we distinguish between \textit{primitive fluents} that are directly affected by operations and \textit{derived fluents} whose truth values are computed from the primitive fluents. 

A \emph{world} $w$ as an instance of the abstract robot system is an assignment of truth values to all predicates. Specifically, $w$ assigns a Boolean value to each rigid predicate instance $P(\vec{o})$ and to each fluent instance $F(\vec{o}, s)$, i.e., $w(P(\vec{o})) \in \{\bot, \top\}$ and $w(F(\vec{o}, s)) \in \{\bot, \top\}$, where $\bot$ and $\top$ denote the Boolean truth values \textit{false} and \textit{true}, respectively. The semantics of formulas follow standard first-order logic. For a formula $\phi$ with a sorted free variable $x$,  $\phi[x/c]$  denotes the formula obtained by substituting every free occurrence of $x$ with  constant $c$. We also denote $D_x$ for the finite domain of variable $x$. The truth of a variable-free formula $\psi$ in world $w$, written $w \models \psi$, is then defined as follows.

\begin{table}[h]
\centering
\vspace{-15pt}
\begin{tabular}{l @{\hspace{1cm}} l}
$w\models P(\vec{o}) \;\text{iff}\; w(P(\vec{o})) = \top$
&
$w\models F(\vec{o},s) \;\text{iff}\; w(F(\vec{o},s)) = \top$
\\

$w\models \psi_1 \land \psi_2 \;\text{iff}\; w\models \psi_1 \ \text{and}\ w\models \psi_2$
&
$w\models \neg\psi \;\text{iff}\; w\not\models \psi$
\\

$w\models \exists x.\phi \;\text{iff}\; w\models \bigvee_{c\in D_x}\phi[x/c]$
&
$w\models \forall x.\phi \;\text{iff}\; w\models \bigwedge_{c\in D_x}\phi[x/c]$
\\
\end{tabular}
\vspace{-15pt}
\end{table}

For a given set of formulas $\Sigma$, we denote $w\models \Sigma$ iff for each $\psi \in \Sigma$, $w\models \psi$.

\paragraph{Action Theories} 
An action theory is a collection of axioms $\mathcal{D}$ that constrain the admissible models of the world, such that any valid world model $w$ satisfies $w \models \mathcal{D}$. The axiom set can be decomposed as $\mathcal{D} = \mathcal{D}_0 \cup \mathcal{D}_p \cup \mathcal{D}_s \cup \mathcal{D}_f$, where \emph{initial state axioms} $\mathcal{D}_0$ specify the properties of the initial situation; 
\emph{precondition axioms} $\mathcal{D}_p$ contain precondition axioms for each operation $\alpha$ of the form $\textsf{Poss}(\alpha,\vec{o},s) \leftrightarrow \psi_\alpha(\vec{o},s)$, where $\psi_\alpha(\vec{o},s)$ is a variable-free formula describing exactly when operation $\alpha$ is accomplishable in situation $s$; \emph{successor state axioms} $\mathcal{D}_s$ characterize the effects of operations on fluents. For each fluent $F$ affected by operation $\alpha$, the corresponding axiom has the general form $\forall \vec o, \alpha, s.~F(\vec{o},do(\alpha,s)) \leftrightarrow \bigl(\gamma^+_{F}(\alpha,\vec{o},s) \lor (F(\vec{o},s) \land \neg \gamma^-_{F}(\alpha,\vec{o},s))\bigr)$, where $\gamma^+_{F}$ specifies the conditions under which $F$ becomes true after executing $\alpha$, and $\gamma^-_{F}$ specifies the conditions under which $F$ becomes false; \emph{foundational axioms} $\mathcal{D}_f$ consists of the foundational axioms of situation calculus that define the structure of situations and the $do$ function which are given in~\cite{lin2008situation}.

\paragraph{Tasks}
The considered situation calculus allows high-level robot tasks to be specified as a program of operations. A task $\tau$ is defined by
\[
\tau ::= \mathbf{nil} \mid \alpha \mid \psi? \mid [\tau_1 ; \tau] \mid [\tau_1 \mid \tau_2]
\]
where $\mathbf{nil}$ is an empty task, $\alpha$ is an operation, $\psi?$ is a test with $\psi$ a variable-free formula, $[\tau_1 ; \tau_2]$ denotes program sequencing, and $[\tau_1 \mid \tau_2]$ denotes nondeterministic choice.  A \textit{task grammar} is then an encapsulation of the above expression, such as ``if $\psi$ then $\tau_1$ else $\tau_2$'' is equivalent to $[[\psi?;\tau_1]\mid [\neg\psi?;\tau_2]]$. For brevity, we also denote $[\alpha_1,\alpha_2,\ldots,\alpha_n] \;\equiv\; [[\ldots[[\alpha_1;\alpha_2],\alpha_3],\ldots],\alpha_n]$ and $[\alpha_1|\alpha_2|\ldots|\alpha_n]\;\equiv\;[[\ldots[[\alpha_1|\alpha_2]|\alpha_3],\ldots]|\alpha_n]$.

Execution of a task is defined by a transition relation over the \textit{execution state} of the form $\langle s,\tau\rangle$, where $s$ is a situation and $\tau$ is a task. For a given world $w$ satisfying the axioms $\mathcal D$, we define the transition relation $\xrightarrow{w}$ over execution states in an inductive manner that mirrors the operational interpretation of task execution. An execution state $\langle s,\tau\rangle$ is considered \emph{final} exactly when $\tau = \mathbf{nil}$, in which case no further transitions are possible. Otherwise, the execution state evolves according to the structure of the task term $\tau$ and the dynamics encoded by $w$, giving rise to the following inductively defined transition rules, where $\phi[s]$ denotes replacing all situation variables in $\phi$ with $s$.
\begin{itemize}
\item $\langle s, \alpha\rangle \xrightarrow{w} \langle do(\alpha,s), \mathbf{nil}\rangle$ if $w \models \textsf{Poss}(\alpha,s)$.
\item $\langle s, \phi?\rangle \xrightarrow{w} \langle s, \mathbf{nil}\rangle$ iff $w \models \phi[s]$.
\item {$\langle s, [\tau_1 ; \tau_2]\rangle \xrightarrow{w} \langle s, \tau_2\rangle$ iff $\tau_1=\mathbf{nil}$.}
\item {$\langle s, [\tau_1 ; \tau_2]\rangle \xrightarrow{w} \langle s', \tau_2\rangle$ iff $\langle s_, \tau_1 \rangle \xrightarrow{w} \langle s', \mathbf{nil}\rangle$.}
\item $\langle s, [\tau_1 \mid \tau_2]\rangle \xrightarrow{w} \langle s', \tau'\rangle$ iff $\langle s, \tau_1\rangle \xrightarrow{w} \langle s', \tau'\rangle$ or $\langle s, \tau_2\rangle \xrightarrow{w} \langle s', \tau'\rangle$.
\end{itemize}

\paragraph{Example.}
We illustrate the modeling process for the motivating example using the introduced situation calculus. It involves four objects: a piece of bread $o_b$, a plate $o_p$, a microwave $o_m$, and a table $o_t$. The operations, rigid predicates, fluents, and axioms for this system are listed in Table~\ref{tab:sc-example}.

The rigid predicates capture invariant properties of objects, providing basic semantic knowledge about them. These include placable relations that specify which objects can be placed on others, as well as predicates indicating whether an object has a door, is heatable, or requires heating. Fluents, in contrast, represent properties that may change across situations as operations are performed. Primitive fluents, such as located, open, or running, can be directly modified by actions like put, open, close, or turn on. In addition to primitive fluents, there may also be derived fluents, whose values are determined from other fluents. For instance, the fluent \textsf{In} can be defined as the transitive closure of located. If a slice of bread is located on a plate and the plate is located in the microwave, then by applying the transitive closure, we infer that the bread is in the microwave.

The set of axioms characterizes the system’s dynamic properties. In particular, the initial-condition axioms specify that no object is running at the outset, and that an object’s open status is fixed as closed whenever it has no door. They also enforce that if two objects do not satisfy the placeable relation, then no located relation can hold between them initially. Moreover, the located relations must be semantically valid, such that each object that is placable on another object must be located on exactly one such object in the initial state.

The precondition axioms and successor-state axioms specify when an operation can be executed and how it affects the primitive fluents. Preconditions range from simple cases, such as open or close, which require that the object has a door, is not already in the corresponding state, and is not running, to more complex ones, like turn on the microwave, which requires that the microwave is closed and that every object inside it is heatable, with at least one object requiring heating. The successor state axioms follow the standard form, specifying how primitive fluents change as a result of an operation. For example, a located fluent between two objects becomes true when a put operation places one object onto or into the other, and becomes false when the object is removed by a put operation that relocates it elsewhere.

\begin{table}[t]
\centering
\caption{The model for the motivating example in situation calculus}
\label{tab:sc-example}
\vspace{-10pt}
\footnotesize
\renewcommand{\arraystretch}{1}

\adjustbox{width=\textwidth}{
\begin{tabular}{p{13cm}}
\hline\\[-3mm]

\textbf{Objects:}
$\mathcal O = \{o_b,o_p,o_m,o_t\}$ \\

\textbf{Operations $\mathcal{A}$:}
$\textsf{put}(o,o'),\; \textsf{open}(o),\; \textsf{close}(o),\; \textsf{turn\_on}(o)$ \\

\textbf{Rigid Predicates:}
$\textsf{Placeable}(o_b, o_m)$,
$\textsf{Placeable}(o_b, o_p)$,
$\textsf{Placeable}(o_p, o_m)$,
$\textsf{Placeable}(o_b, o_t)$,
$\textsf{Placeable}(o_p, o_t)$,
$\textsf{HasDoor}(o_m)$,
$\textsf{Microwave}(o_m)$,
$\textsf{Heatable}(o_b)$,
$\textsf{Heatable}(o_p)$,
$\textsf{RequireHeat}(o_b)$ \\

\textbf{Primitive fluents}:
$\textsf{Loc}(o,o',s),\; \textsf{IsOpen}(o,s),\; \textsf{Running}(o,s)$ \\

\textbf{Derived fluents}:
$\textsf{In}(o,o',s)$ (transitive closure of $\textsf{Loc}$) \\

\textbf{Initial condition axioms $\mathcal D_0$:}
\begin{itemize}[itemsep=0pt, topsep=0pt, partopsep=0pt, parsep=0pt, leftmargin=1em]
\item $\forall o.\; \neg \textsf{Running}(o,s_0)$,\;
\item $\forall o.\; \neg \textsf{HasDoor}(o)\to  \textsf{IsOpen}(o, s_0)$,\;
\item $\forall o.\forall o'.\; \neg \textsf{Placeable}(o,o')\to \neg \textsf{Loc}(o,o',s_0)$ 
\item $\forall o. \forall o'. \forall o''.\;((\textsf{Loc}(o, o', s_0)\land o'\neq o'')\to \lnot \textsf{Loc}(o, o'', s_0))$
\item $\forall o. (\exists o'. \textsf{Placeable}(o, o')) \to (\exists o'. \textsf{Loc}(o, o', s_0))$
\end{itemize}\\

\textbf{Precondition axioms $\mathcal D_p$:}
\begin{itemize}[itemsep=0pt, topsep=0pt, partopsep=0pt, parsep=0pt, leftmargin=1em]
    \item $\forall o.\;\textsf{Poss}(\textsf{open}(o),s)\leftrightarrow (\textsf{HasDoor}(o)\wedge \neg \textsf{IsOpen}(o,s)\wedge \neg \textsf{Running}(o,s))$
    
    \item $\forall o.\;\textsf{Poss}(\textsf{close}(o),s)\leftrightarrow (\textsf{HasDoor}(o)\wedge \textsf{IsOpen}(o_m,s)\wedge \neg \textsf{Running}(o_m,s))$
    
    \item $\forall o.\;\textsf{Poss}(\textsf{turn\_on}(o),s)\leftrightarrow (\textsf{Microwave}(o)\wedge \neg \textsf{IsOpen}(o,s)\wedge \forall o'.(\textsf{In}(o',o,s)\to \textsf{Heatable}(o'))\wedge \exists o.\;(\textsf{In}(o,o_m,s)\wedge \textsf{RequireHeat}(o)))$
    
    \item $\forall o.\forall o'.\;\textsf{Poss}(\textsf{put}(o,o')) \leftrightarrow (\textsf{Placeable}(o,o') \wedge \lnot \textsf{In}(o, o') \wedge (\textsf{HasDoor}(o')\to \textsf{IsOpen}(o')) \wedge (\forall o''.\; (\textsf{In}(o,o'')\wedge \textsf{HasDoor}(o''))\to \textsf{IsOpen}(o'')))$
\end{itemize}\\
\textbf{Successor state axioms $\mathcal D_s$:}
\begin{itemize}[itemsep=0pt, topsep=0pt, partopsep=0pt, parsep=0pt, leftmargin=1em]

    \item $\forall o. \forall o'. \forall \alpha.\;\textsf{Loc}(o,o',\textsf{do}(\alpha,s)) \leftrightarrow 
    \gamma^+_{\textsf{Loc}}(\alpha,o,o',s)\;\lor\;
    (\textsf{Loc}(o,o',s)\wedge \neg \gamma^-_{\textsf{Loc}}(\alpha,o,o',s))$, 

    where  
    $\gamma^+_{\textsf{Loc}}(\alpha,o,o',s)\leftrightarrow (\alpha=\textsf{put}(o,o'))$  
    and  
    $\gamma^-_{\textsf{Loc}}(\alpha,o,o',s)\leftrightarrow \exists o''.\,(\alpha=\textsf{put}(o,o''))$.
    
    \item $\forall o.\forall \alpha.\;\textsf{IsOpen}(o,\textsf{do}(\alpha,s)) \leftrightarrow 
    \gamma^+_{\textsf{IsOpen}}(\alpha,o,s)\;\lor\;
    (\textsf{IsOpen}(o,s)\wedge \neg \gamma^-_{\textsf{IsOpen}}(\alpha,o',s))$,

    where  
    $\gamma^+_{\textsf{IsOpen}}(\alpha,o,s)\leftrightarrow (\alpha=\textsf{open}(o))$  
    and  
    $\gamma^-_{\textsf{IsOpen}}(\alpha,o,s)\leftrightarrow (\alpha=\textsf{close}(o))$.

    \item $\forall o. \forall \alpha.\;\textsf{Running}(o,\textsf{do}(\alpha,s)) \leftrightarrow 
    \gamma^+_{\textsf{Running}}(\alpha,o',s)\;\lor\;
    (\textsf{Running}(o',s)\wedge \neg \gamma^-_{\textsf{Running}}(\alpha,o',s))$,

    where  
    $\gamma^+_{\textsf{Running}}(\alpha,o',s)\leftrightarrow (\alpha=\textsf{turn\_on}(o'))$  
    and  
    $\gamma^-_{\textsf{Running}}(\alpha,o',s)\leftrightarrow \textsf{false}$.
\end{itemize}\\
\hline
\end{tabular}
}
\end{table}

A compound task such as ``put $o_b$ into $o_m$ and then close the door of $o_m$'' is formulated as $
  [\,\textsf{put}(o_b,o_m)\,;\ \textsf{close}(o_m)\,].
$ A conditional task such as ``if the door of $o_m$ is closed, open it'' is expressed as $[\,\neg \textsf{IsOpen}(o_m,o_s)?\ ;\ \textsf{open}(o_m)\,]$. Given a world $w$ satisfying the axioms $\mathcal{D}$, task execution proceeds by transitions of the form $\langle s,\tau\rangle \xrightarrow{w} \langle s',\tau'\rangle$, as defined earlier. For example, if the door is initially closed (i.e., $w \models \neg \textsf{IsOpen}(o_m,o_s)$), executing the above conditional task yields the transition
$\langle s_0,\ [\neg \textsf{IsOpen}(o_m,o_s)?;\ \textsf{open}(o_m)] \rangle
  \xrightarrow{w}
\langle s_0,\ \textsf{open}(o_m) \rangle
  \xrightarrow{w}
\langle \textsf{do}(\textsf{open}(o_m),s_0),\ \mathbf{nil} \rangle.
$

\subsection{Constructing Constraint-Aware Combinatorial Testing Model}

By adopting situation calculus for modeling a robot system, one can capture the complex interactions between operations and objects as long as the relevant objects and predicates remain finite. With this assumption, the {world state} of the initial situation $s_0$ can be characterized by the truth assignment over all fluents that refer to $s_0$. We denote this initial world state by $w_0$, which is the fragment of the full world model $w$ restricted to fluents whose situation is $s_0$. Because both objects and predicate schemas are finite, the space of all possible $w_0$ is also finite and can be exhaustively enumerated.

Validation requires comprehensively exploring both initial system states and tasks, while ensuring that each task is accomplishable from its corresponding initial state. For practical reasons, we restrict our attention to tasks whose derivation length is at most $K$, that is, the task can be generated by expanding its syntax within $K$ steps, making the overall task space finite. To determine accomplishability, we first enumerate candidate tasks and calculate the task-level precondition for each, which must hold in the initial world {state} for the task to be accomplishable. Thus, a task is considered accomplishable for an initial world {state} $w_0$ only if $w_0$ satisfies this task-level precondition.  Finally, we encode these constraints into a combinatorial model whose parameters jointly represent the initial world state and the task structure, enabling the generation of \emph{configurations} of $( w_0, \tau)$ that are guaranteed to satisfy the accomplishability requirement.

\subsubsection{Regression-Based Computation of Weakest Preconditions}
In this section, we demonstrate how to propagate task success conditions backward through the task structure using regression-based computation. Let $\mathcal{D} = \mathcal{D}_0 \cup \mathcal{D}_p,\mathcal{D}_s \cup \mathcal{D}_f$ be the axioms of the system.

Regression is a syntactic transformation, denoted by $\textsf{Regr}$, that translates a formula $\psi$ holding in a successor situation $do(\alpha, s)$ into a logically equivalent formula holding in the current situation $s$, relative to the set of successor state axioms $\mathcal{D}_s$. Specifically, for a primitive fluent $F$ governed by $\mathcal{D}_s$, the regression is defined by substituting the fluent with the right-hand side of its axiom:$
\textsf{Regr}(F(\vec{o}, do(\alpha,s))) \;\stackrel{\text{def}}{=}\; \gamma^+_{F}(\alpha,\vec{o},s) \,\lor\, \bigl(F(\vec{o},s) \land \neg \gamma^-_{F}(\alpha,\vec{o},s)\bigr).
$
This transformation is extended to arbitrary formulas recursively. Since operation $\alpha$ is embedded within the situation term of the fluents, the recursive rules propagate the operator: $\textsf{Regr}(\neg \phi) = \neg \textsf{Regr}(\phi),$ $\textsf{Regr}(\phi_1 \land \phi_2) = \textsf{Regr}(\phi_1) \land \textsf{Regr}(\phi_2)$.

Based on the regression computation, we introduce a weakest-precondition operator $\textsf{WP}(\varphi,\tau)$ to determine whether a task $\tau$ is accomplishable and achieve a desired postcondition $\varphi$ in situation $s$. Intuitively, $\mathit{WP}(\varphi,\tau)$ characterizes exactly those initial situations from which executing $\tau$ is both possible and guarantees that $\varphi$ will hold afterward. This is done by jointly using the $Regr$ of the post condition and $Poss$ for the applied operation. Regression $\textsf{Regr}$ normally operates on a formula whose situation term is already embedded in the formula. In contrast, $\mathit{WP}$ takes as input a goal formula $\varphi(s)$ that explicitly mentions a situation variable $s$. Therefore, when computing the weakest precondition for a primitive operation, we substitute the occurrence of $s$ in $\varphi$ with the successor situation $do(\alpha,s)$, and then apply one step of the regression transformation. This yields a condition that must hold before performing operation $\alpha$.
$$
\mathit{WP}(\varphi, \alpha) = \textsf{Poss}(\alpha, \vec o, s) \land \textsf{Regr}\bigl(\varphi[s/ do(\alpha,s)]\bigr).
$$
Then, for any task, its weakest precondition is computed inductively as follows:
\begin{itemize}
    \item $\mathit{WP}(\varphi, \mathbf{nil}) = \varphi$.
   
    \item $\mathit{WP}(\varphi, \psi?) = \phi \land \psi$.
    
    \item $\mathit{WP}(\varphi, [\tau_1 ; \tau_2]) = \mathit{WP}\bigl(\mathit{WP}(\varphi, \tau_2), \tau_1\bigr)$.
    
    \item $\mathit{WP}(\varphi, [t_1 \mid t_2]) = \mathit{WP}(\varphi, t_1) \,\lor\, \mathit{WP}(\varphi, t_2)$.
\end{itemize}

\begin{table}[t]
\caption{Examples for computing weakest preconditions for tasks}
\label{tab:task}
\vspace{-5pt}
\adjustbox{width=\textwidth}{
\begin{tabular}{p{7.2cm} p{7.2cm}}
\hline
\textbf{Task 1: $[\textsf{open}(o_m);\;\textsf{close}(o_m)]$}
&
\textbf{Task 2: $[\textsf{open}(o_m);\;\textsf{open}(o_m)]$}
\\
\hline

$
\begin{array}{l}
\mathit{WP}([\textsf{open}(o_m);\textsf{close}(o_m)],\top) \\
\;\;= WP(WP(\top,\textsf{close}(o_m)),\textsf{open}(o_m)) \\[4pt]
\mathit{WP}(\top,\textsf{close}(o_m)) \\
\;\;= \textsf{IsOpen}(o_m,s)\wedge \neg\textsf{Running}(o_m,s) \\[4pt]
\mathit{WP}(\textsf{IsOpen}(o_m,s)\wedge\neg\textsf{Running}(o_m,s),\textsf{open}(o_m)) \\
\;\;= \neg\textsf{IsOpen}(o_m,s)\wedge \neg\textsf{Running}(o_m,s)
\end{array}
$
&
$
\begin{array}{l}
\mathit{WP}([\textsf{open}(o_m);\textsf{open}(o_m)],\top) \\
\;\;= WP(WP(\top,\textsf{open}(o_m)),\textsf{open}(o_m)) \\[4pt]
\mathit{WP}(\top,\textsf{open}(o_m)) \\
\;\;= \neg\textsf{IsOpen}(o_m,s)\wedge \neg\textsf{Running}(o_m,s) \\[4pt]
\mathit{WP}(\neg\textsf{IsOpen}(o_m,s)\wedge\neg\textsf{Running}(o_m,s),\textsf{open}(o_m)) \\
\;\;= \bot
\end{array}
$
\\
\hline
\end{tabular}
}
\end{table}

\paragraph{Example.}

We present weakest-precondition reasoning on two tasks for the abstract robot system in Table~\ref{tab:task}: $[\textsf{open}(o_m);\textsf{close}(o_m)]$ and $[\textsf{open}(o_m);\textsf{open}(o_m)]$. As shown in the table, the first task is accomplishable only when the microwave door is initially closed, and the microwave is not running. In contrast, the weakest precondition of the second task is \(\bot\), indicating that no initial state can satisfy the requirements for executing two consecutive \textsf{open} operations. Hence, Task~2 is unaccomplishable.

\subsubsection{Constraint-Aware Combinatorial Test Construction}

Building a combinatorial testing model for generating the world-task configuration relies on a given abstract robot system model with objects $\mathcal O$, operations $\mathcal{A}$, fluents $\mathcal{F}$, and the axioms $\mathcal D$. In addition, we assume a set of task grammar rules $\mathcal G_r$ is given, where each rule has the form {$X ::= e$}, where $X$ is a nonterminal and $e$ is an expression, such as $\tau ::= \textsf{if}\ \psi\ \textsf{then}\ \tau_1\ \textsf{else}\ \tau_2$; or $\tau::=\alpha$.  A task can be generated by repeatedly expanding the leftmost nonterminal until only terminal symbols remain.

To encode the initial world state as the fluent assignments in the combinatorial model, we distinguish between unary fluents, which take a single object argument, and multi-ary fluents, whose arity is at least two.
For a unary fluent $F(o, s_0)$ with one object argument, we simply introduce a boolean parameter in the combinatorial testing model,
$F_o \in \{\bot, \top\}$, indicating whether $F(o,s_0)$ holds in the initial situation. 
For a $n$-ary fluent $F(o_1,\ldots,o_n, s_0)$ with $n \ge 2$, we consider a finite number of instances of them not exceeding an upper bound derived from the initial-state axioms. In the worst case, if no constraint is applicable, the number of possible instances is bounded by $|\mathcal O|^n$. In the combinatorial model, each possible instance is represented by a parameter tuple $(F_{i,1},\ldots,F_{i,n}) \in {\mathcal O}^n \cup \{\varepsilon\}^n$. 
{The fluent $F(o_1,\ldots,o_n,s_0)$ holds iff there exists an index $i$ such that $F_{i,1}=o_1,\ldots,F_{i,n}=o_n$, and $(\varepsilon, ..., \varepsilon)$ denotes there is no fluent term is assigned to be true according to these parameters.} To eliminate symmetric assignments, we impose a total order $<_o$ over $\mathcal O^n \cup \{\varepsilon\}^n$, extend it lexicographically to tuples, and regard $(\varepsilon,\ldots,\varepsilon)$ as the maximum element. We then require $(F_{i,1},\ldots,F_{i,n}) <_o (F_{i+1,1},\ldots,F_{i+1,n})$ and allow equalitiy only when $(F_{i,1}, ..., F_{i,n})= (\varepsilon, ...\varepsilon)$, ensuring that instance $i+1$ encodes a strictly larger tuple than instance $i$ and thereby enforcing a canonical representation of fluent assignments. Finally, constraints induced by the initial condition axioms $\mathcal{D}_0$ are encoded into the model by replacing the predicates with the conjunction of assignments in the combinatorial testing model.

To encode task generation within the same combinatorial testing framework while enforcing a derivation-depth bound \(K\), each derivation step is modeled as a parameter $
d_k \in G_r \cup \{\epsilon\},
$ where $1\le k\le K$ and \(\epsilon\) denotes that no further expansion is applied. A set of constraints ensures that any chosen sequence of steps constitutes a valid derivation and eliminates symmetric variants, following the standard grammar-based combinatorial construction of~\cite{hoffman2011grammar}. Enumerating all valid assignments \((d_1,\dots,d_K)\) yields a finite collection of syntactically correct tasks. For each generated task \(\tau\), we then compute its weakest precondition \(\mathit{WP}(\tau)\), which specifies the preconditions over the initial world states from which the task is accomplishable. Finally, these preconditions are added as constraints in the form $(d_1=r_1\wedge d_2=r_2\wedge ... \wedge d_n=r_n)\to \mathit{WP}(\tau[r_1,r_2,...,r_n])$, where $r_1,...,r_n\in \mathcal G_r$ are grammar rules, and $\tau[r_1,r_2,...,r_n]$ is the task generated by successively applying the rules $r_1,...,r_n$ to the top nonterminal $\tau$.

Solving the combinatorial testing model yields valid world–task configurations. This model encodes fluent assignments for the initial situation, the stepwise application of grammar rules for task generation, and all associated constraints. Each parameter assignmen corresponds exactly to a configuration where the generated task is accomplishable from its specified initial world state.

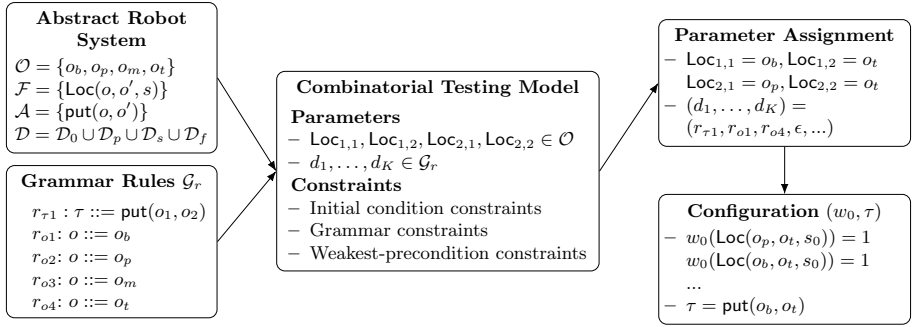
\begin{figure}[t]
\centering
\adjustbox{width=\textwidth}{
\begin{tikzpicture}[
  >=Latex,
  node distance=8mm and 20mm,
  block/.style={draw, rounded corners, align=center, inner sep=4pt, minimum width=3.2cm},
  smallblock/.style={draw, rounded corners, align=left, inner sep=3pt, minimum width=3.6cm, font=\footnotesize}
]

\node[block] (grammar) {
    \textbf{Grammar Rules $\mathcal{G}_r$}\\[2mm]
\hspace{0.3cm}\begin{minipage}{3.0cm}
    \raggedright
    $r_{\tau1}: \tau ::= \textsf{put}(o_1,o_2)$    $r_{o1}$: $o ::= o_b$ \\
    $r_{o2}$: $o ::= o_p$ \\
    $r_{o3}$: $o ::= o_m$ \\
    $r_{o4}$: $o ::= o_t$
    \end{minipage}
};
\node[block, above=2 mm of grammar] (fluents) {
\textbf{Abstract Robot}\\
\textbf{System} \\[1mm]
\begin{minipage}{3.3cm}
$\mathcal O =\{o_b,o_p,o_m,o_t\}$\\
$\mathcal F =\{\textsf{Loc}(o,o',s)\}$\\
$\mathcal A =\{\textsf{put}(o, o')\}$\\
$\mathcal D = \mathcal D_0 \cup \mathcal D_p \cup \mathcal D_s \cup \mathcal D_f$
\end{minipage}
};

\node[block, right=10mm of grammar, yshift=1.2cm, minimum height=26mm] (ctmodel) {
\textbf{Combinatorial Testing Model}\\[2mm]
\hspace{1mm}\begin{minipage}{5.1cm}
    \raggedright
  \textbf{Parameters}\\
  \begin{itemize}[itemsep=0pt, topsep=0pt, partopsep=0pt, parsep=0pt, leftmargin=1em]
  	\item $\textsf{Loc}_{1,1}, \textsf{Loc}_{1,2}, \textsf{Loc}_{2,1}, \textsf{Loc}_{2,2}\in \mathcal O$
  	\item $d_1,\dots,d_K\in \mathcal G_r$

  \end{itemize}

  \textbf{Constraints}\\
  \begin{itemize}[itemsep=0pt, topsep=0pt, partopsep=0pt, parsep=0pt, leftmargin=1em]
    \item Initial condition constraints
  	
  	\item Grammar constraints
  	
  	\item Weakest-precondition constraints

  \end{itemize}

\end{minipage}
};

\node[block, right=10mm of ctmodel, yshift=1.5cm] (config) {
\textbf{Parameter Assignment}\\[1mm]
\begin{minipage}{4cm}
  \begin{itemize}[itemsep=0pt, topsep=0pt, partopsep=0pt, parsep=0pt, leftmargin=1em]
  \item $\textsf{Loc}_{1,1} = o_b, \textsf{Loc}_{1,2}=o_t$
  
  $\textsf{Loc}_{2,1} = o_p, \textsf{Loc}_{2,2}=o_t$
  
   \item   $(d_1,\dots,d_K)=$
   
   $(r_{\tau1}, r_{o1}, r_{o4}, \epsilon, ...)$

  \end{itemize}
 \end{minipage}
};

\node[block, right=10mm of ctmodel, yshift=-1.5cm] (wt) {
\textbf{Configuration} $(w_0, \tau)$\\[1mm]
\begin{minipage}{4cm}
  \begin{itemize}[itemsep=0pt, topsep=0pt, partopsep=0pt, parsep=0pt, leftmargin=1em]
	\item $w_0(\textsf{Loc}(o_p, o_t, s_0)) = 1$	

          $w_0(\textsf{Loc}(o_b, o_t, s_0)) = 1$
          
            	...
    
    \item $\tau=\textsf{put}(o_b, o_t)$
  \end{itemize}

\end{minipage}
};

\draw[->] (grammar.east) -- (ctmodel.west);
\draw[->] (fluents.east) -- (ctmodel.west);

\draw[->] (ctmodel.east) -- (config.west);
\draw[->] (config.south) -- (wt);

\end{tikzpicture}
}

\caption{From grammar and abstract robot system to a constraint-aware combinatorial testing model and configurations.}
\label{fig:ct-example}
\vspace{-3mm}
\end{figure}

\paragraph{Example} Consider the system in Fig.~\ref{fig:ct-example}, which extracts the fragment of the abstract model in Table~\ref{tab:sc-example} relevant to the \textsf{put} operation. We also assume a simplified grammar $\mathcal G_r$, where the task reduces to a single  \textsf{put} operation. 

From the initial state axioms, exactly two \textsf{Loc} fluents hold, since only bread and plate are placeable and both of them must be located on another. After breaking the symmetry, we obtain $\textsf{Loc}_{1,1}=o_b$ and $\textsf{Loc}_{2,1}=o_p$, leaving $\textsf{Loc}_{1,2}$ and $\textsf{Loc}_{1,2}$ to be chosen from their domains. The combinatorial testing model further contains parameters that record the grammar rules sequentially applied to the root nonterminal $\tau$. For example, a derivation may proceed as {$\tau \Rightarrow_{r_{\tau1}} \textsf{put}(o_1,o_2) \Rightarrow_{r_{o1}} \textsf{put}(o_b,o_2) \Rightarrow_{r_{o4}} \textsf{put}(o_b,o_t)$}, where $\Rightarrow_r$ denotes the application of the grammar rule $r$ to the leftmost nonterminal. 

In this example, $d_1$ has only one possible value, $r_{\tau1}$, whereas different choices for $d_2$ and $d_3$ yield different \textsf{put} tasks. Similarly, varying the assignments of $\textsf{Loc}_{1,2}$ and $\textsf{Loc}_{2,2}$ results in different initial fluent assignments. A parameter assigment in the combinatorial testing model therefore corresponds to a triplet $(o_1,o_2,o_3)$ appearing in $\textsf{Loc}(o_1,o_2)$ and $\textsf{put}(o_1,o_3)$, yielding combinations such as $({o}_{b},{o}_{p}, {o}_{m})$, $(o_b,o_m,o_p)$, and so on, for a total of eight valid combinations. As the number of objects or the complexity of grammar increases, the number of potential combinations grows exponentially. However, combinatorial testing reduces the test budget by controlling the coverage strength: for 1-way coverage, only three assignments, such as $(o_b,o_m,o_p)$, $(o_p,o_t,o_m)$, and $(o_b,o_p,o_t)$, are sufficient, giving that every parameter value appears in at least one assigment.

\subsection{Falsifying Concrete Robot System via Mapping the Abstract Configuration}

In this section, we describe how abstract configurations are mapped to concrete robot systems for falsification. We first synthesize STL specifications whose predicates correspond to signals observable in the concrete system, and then present the falsification procedure through optimization.

\subsubsection{From Tasks to STL Specifications}

Recall that in a world–task configuration, tasks are defined as programs composed of operations. Executing a task from the initial world state determines the resulting situation. Since a situation represents a history of operations, it is possible to infer the fluent assignments at every step, starting from the initial state and continuing after each operation, until the task is complete. Accordingly, we define task completion for the concrete system as the requirement that its state sequence conforms to the fluent evolution dictated by the corresponding logical situation. We assume the states of the concrete system are fully observable and are timed signals such that a world-task configuration can be translated to an STL formula as follows.

We first construct a mapping from fluent assignments to predicates over the observable signals of the concrete system. This is feasible in simulators with full observability. For example, the fluent $\textsf{Open}(o,s)$ can be evaluated by checking whether the signal corresponding to the door angle of $o$ exceeds a threshold, e.g., $\textsf{DoorAngle}(o)>80^\circ$, and $\textsf{Loc}(o,o’,s)$ can be evaluated by checking whether the distance between the surfaces of $o$ and $o’$ is sufficiently close, such as $\textsf{dist}(o_1, o_2)\le 0.01$ m. With these predicates in place, we can relate fluent assignments in the abstract world to observations of the concrete system.  For each fluent instance $F(\vec o,s)$, we introduce a predicate $p_{F,\vec o}$ over the concrete signals. The abstract world state $w$ at situation $s$ is then mapped to the concrete system by the propositional formula
$
\chi_{w,s} \;\stackrel{\text{def}}{=}\;
(
  \bigwedge_{F,\vec o :\, w \models F(\vec o,s)} p_{F,\vec o}
)
\;\wedge\;
(
  \bigwedge_{F,\vec o :\, w \models \neg F(\vec o,s)} \neg p_{F,\vec o}
)$.

We now consider a task given as a sequence of primitive operations $[\alpha_0;\alpha_1;\ldots;$ $\alpha_n]$ executed from a known initial world state $w_0$. For the induced situations $s_{i+1}=do(\alpha_i,s_i)$, the fluent assignment $w_{i+1}$ can be infered by the assignment $w_i$ for the previous situation $s_i$ using the successor state axioms, such that  $w_{i+1}\models F(\vec o,s_{i+1})$ holds iff (i) $w_i\models \gamma^{+}(F,\alpha_i,\vec o,s_i)$, (ii) $w_i\models \neg \gamma^{-}(F,\alpha_i,\vec o,s_i)$, or (iii) $w_i\models F(\vec o,s_i)$ when the operation does not affect the fluent. With the fluent assignment $w_i$ known for each situation $s_i$, we induce a propositional requirement over the STL observation predicates. Assuming each primitive operation completes within a duration $\Delta t$, the STL specification for the entire sequence is defined inductively as $\textsf{STL}([\alpha_0;\ldots;\alpha_n]) = \Diamond_{[0,\Delta t]}(\chi_{w_1, s_1} \wedge \Diamond_{[0,\Delta t]}(\chi_{w_2,s_2} \wedge \cdots \Diamond_{[0,\Delta t]} \chi_{w_n, s_n}))$, expressing that after each operation, the concrete execution must satisfy the corresponding fluent assignment within the allowed time window.

For tasks containing nondeterministic choice $\tau \mid \tau'$ and test constructs $\psi?$, we first rewrite them using associativity and distributivity, namely $[[\tau \mid \tau'];\tau''] \equiv [[\tau;\tau''] \mid [\tau';\tau'']]$ and $[\tau'';[\tau \mid \tau']] \equiv [[\tau'';\tau] \mid [\tau'';\tau']]$, yielding a normal form $\tau_1 \mid \cdots \mid \tau_m$ in which each branch $\tau_i$ is a sequence of primitive operations or tests. We then prune any branch $\tau_i$ for which $w_0\not\models WP(\tau_i)$, and within each remaining branch remove its test terms, 
since $w_0\models WP(\tau_i)$ guarantees their satisfaction. This produces a nondeterministic set of operation-only sequences accomplishable from $w_0$. The STL specification of the nondeterministic task is therefore the disjunction over its accomplishable branches: $\textsf{STL}(\tau_1 \mid \cdots \mid \tau_m) = \textsf{STL}(\tau_1) \vee \cdots \vee \textsf{STL}(\tau_m)$.

\subsubsection{Falsification problem}

Given a generated configuration $(w_0,\tau)$ and the corresponding specification $\mathsf{STL}(\tau)$, falsification proceeds by instantiating the abstract initial world state on the concrete robot system. This is done by reversing the fluent-to-signal mapping introduced earlier to obtain a parameter space $Q_{w_0}$ of concrete initial states consistent with the fluent assignment $w_0$. Executing the task $\tau$ from any $q_0\in Q_{w_0}$ under the robot policy yields a concrete trajectory $pol(q_0,\tau)=q_1,q_2,\ldots,q_n$. This finite sequence of concrete state signals can then be evaluated against the STL specification to determine whether the execution violates $\mathsf{STL}(\tau)$. Then the falsification problem can be modeld as the following optimization problem:
\begin{align*}
\min_{q_0\in Q_{w_0}}
\ \rho\bigl(\mathsf{STL}(\tau),\, pol(q_0, \tau),\,0\bigr)
\quad
\end{align*}
where the robustness value $\rho(\mathsf{STL}(\tau), q, 0)$ measures how well the execution $q$ satisfies $\mathsf{STL}(\tau)$ at time~0. A negative optimum indicates a violation, yielding a concrete counterexample.

\section{Evaluation}

We have implemented the proposed validation framework by extending IndiGolog~\cite{de2009indigolog} to support situation-calculus-based reasoning over tasks. Combinatorial testing is realized through a greedy selection strategy~\cite{nie2011survey}. Falsification is performed by integrating an optimizer (Nevergrad~\cite{bennet2021nevergrad}) with an STL monitor (RTAMT~\cite{nivckovic2020rtamt}). Finally, we connect the framework with the simulated robot system using the NVIDIA GR00T-N1.5~\cite{bjorck2025gr00t} controller and the RoboCasa environment~\cite{nasiriany2024robocasa} and extend the model presented in Table~\ref{tab:sc-example} to 8 objects. We evaluate the framework by addressing the following research questions (RQs):
\begin{itemize}
    \item \textbf{RQ1}: How effectively does the framework generate valid world-task configurations?
    \item \textbf{RQ2}: How effectively are counterexamples found for the robot system?
\end{itemize}
\subsubsection{Answer to RQ1}

With the goal of exploring possible world-task configurations within a bounded task-syntax depth, and ensuring controlled coverage via combinatorial testing, we summarize the resulting scale of tasks and world-task combinations in Table~\ref{tab:summary}.

\begin{table}[h]
\centering
\caption{Number of world-task configurations and their t-way coverage}
\label{tab:summary}
\vspace{-5pt}
\setlength{\tabcolsep}{4pt}
\begin{tabular}{c|cc|c|ccc}
\hline
\multirow{2}{*}{\begin{tabular}{c}
	Syntax\\Depth
\end{tabular}} & 
\multicolumn{2}{c|}{Task}&\multicolumn{4}{c}{world-task configuration}
 \\
\cline{2-7}
& Syntax valid & Accomplishable& Full coverage &
1-way & 2-way & 3-way \\
\hline
4  & 27   & 25   & 25  &  9 &  25 &  25 \\
6  & 63   & 25   & 25  &  9 &  25 &  25 \\
8  & 583  & 277  & 523 & 23 & 171 & 407 \\
10 & 4279 & 345  & 773 & 26 & 224 & 637 \\
\hline
\end{tabular}
\end{table}

The table reports the number of syntactically valid tasks generated by applying the production rules up to a given depth. As expected, this number grows exponentially: from only 27 tasks at depth 4, to 4,279 tasks when the depth increases to 10. However, only a small fraction of these syntactically valid tasks are accomplishable with satisfiable initial world state conditions. In particular, among the 4,279 syntactically valid tasks at depth 10, only 345 are accomplishable, accounting for just 8.1\%. This highlights the necessity of semantic analysis using situation calculus: it effectively filters out 91.9\% of grammar-generated tasks that are syntactically correct but unaccomplishable for any initial conditions.

When analyzing the generated world-task configurations, we observe that the number of configurations is often roughly twice the number of accomplishable tasks at higher syntax depths. This occurs because a single task-level weakest precondition may be satisfied by multiple distinct assignments to the initial fluents. The total number of configurations can be substantially reduced when using 1-way or 2-way coverage, since these settings require fewer combinations of object attributes to be exercised. In contrast, the reduction is less significant when full coverage already yields a small configuration space, or when the coverage strength exceeds three, as higher-order interactions dominate the combinatorial structure.

\begin{figure}[t]
	\center
	\includegraphics[width=\textwidth]{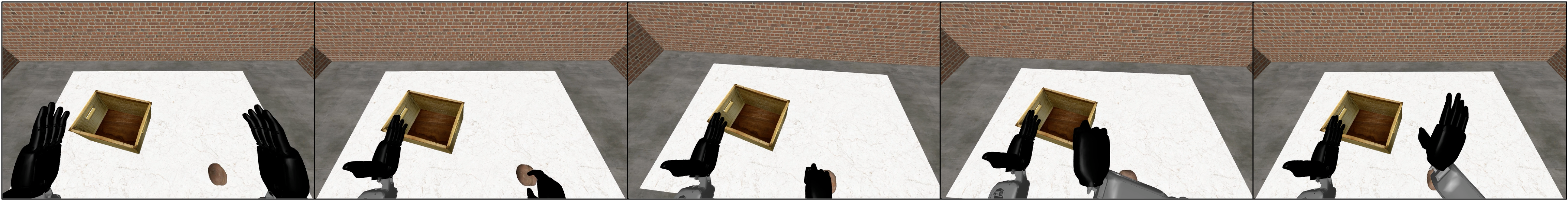}

	\center
	\includegraphics[width=\textwidth]{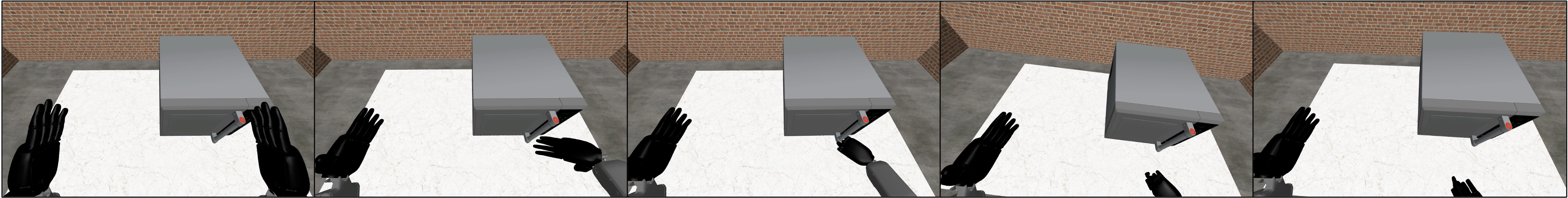}
	\vspace{-15pt}
	\caption{Some of the uncovered counterexamples}
	\label{fig:counterexample}
    
\end{figure}
\subsubsection{Answer to RQ2}

As the generalist robot system is currently in an early stage of development, we focus on simple tasks by setting the depth boundary to 4. With this setting, we perform full-coverage validation over all valid world-task configurations. Our validation framework generates {25} world-task configurations. For each configuration, the falsification process is allowed to produce up to five scenarios. Only {3} configurations passed the validation, meaning no counterexample was found that violates the task specification. The remaining {22} configurations were falsified in a single iteration. These results indicate that there remains a substantial gap between the current state of the art and the actual requirements for a generalist robot. Fig.~\ref{fig:counterexample} presents two counterexamples for GR00T: 1) a failure to pick up an apple for the task ``\emph{put the apple into the drawer}''; and 2) a failure to open a microwave for the task ``\emph{open the microwave's door}''.

\section{Related Work}

Validation and testing of autonomous agents have been explored through a wide range of methodologies, including formal verification, model-based testing, simulation-based falsification, and data-driven evaluation. Recent systematic reviews provide comprehensive coverage of this landscape~\cite{araujo2023testing,tang2023survey}. However, these methods (as mentioned in Sec.~\ref{sec:introduction}) are not adaptable to general-purpose robots, given that the operational domains and specifications cannot be assumed in advance.

Rather than relying on a fixed operational domain, we analyze the primitive components that constitute robot tasks and environments, namely primitive operations, object types, and fluents describing evolving situations. This contrasts with approaches that rely on data-driven coverage criteria~\cite{li2023behavior}. Moreover, we do not use these primitives to model or approximate the controller's policy. Instead, they are used to verify that the generated world–task configurations are accomplishable and reflect the behaviors expected of a general-purpose robot. Validation, therefore, boils down to falsifying the system under test against these expected behaviors. In this sense, our approach differs from~\cite{aineto2023falsification}, which also adopts a language for worlds as we do, but relies on an explicit model of the controller's behavior.

We support flexible and sound reasoning by adopting the situation calculus, a formalism widely used to model dynamic robotic systems. Prior work using the situation calculus has largely focused on planning, task execution, or activity recognition~\cite{hofmann2025ltlf}. In contrast, our method performs systematic exploration of system evolution through bounded, coverage-controlled generation using combinatorial testing. Other modeling formalisms, such as STRIPS~\cite{fikes1971strips,suarez2020strips}, ADL~\cite{pednault1989adl}, and PDDL~\cite{aeronautiques1998pddl}, are also used in robotics. Our two-layer validation framework is bound to situation calculus and can also be applied in other robot modelling languages. However, these languages are generally less expressive than the situation calculus and may provide a limited foundation for complex reasoning~\cite{roger2007expressiveness}.

\section{Conclusion}

This work represented a first step toward a systematic and rigorous validation framework for generalist robots. We built a compact symbolic model based on a small set of primitive operations, a collection of objects, and fluents representing object states and relations. By combining these primitives with a coverage-driven generation of world-task configurations, our method supports flexible creation of hundreds or even thousands of distinct configurations. Through evaluation, we have shown that each generated configuration can be used for falsification-based validation of robot behavior under diverse conditions.

Since the proposed modeling and validation techniques are general, the framework can easily be adapted to applications beyond the tabletop domain. Future work could involve expanding the vocabulary of objects and operations to cover more complex tasks, and linking to training pipelines or higher-level robotic software stacks, leveraging the framework's strong data-synthesis capabilities. Yet another interesting direction is to move beyond a black-box validation and testing setting by extracting the internal signals of the vision-language-action model to isolate states where eventual failure is unavoidable.

\bibliographystyle{splncs04}
\bibliography{reference.bib}

\end{document}